\documentclass[10pt,twocolumn,letterpaper]{article}

\usepackage{ABCD}
\usepackage{times}
\usepackage{epsfig}
\usepackage{graphicx}
\usepackage{amsmath}
\usepackage{amssymb}
\usepackage{enumitem}
\usepackage{subcaption}
\usepackage{dsfont}
\usepackage{pifont}
\usepackage{blindtext}
\usepackage{morewrites}

\usepackage[pagebackref=true,breaklinks=true,letterpaper=true,colorlinks,bookmarks=false]{hyperref}

\newcommand{\xmark}{\ding{55}}%
\newcommand{\cmark}{\ding{51}}%

\ABCDfinalcopy

\ifABCDfinal\fi

\begin{document}

\title{Temporal Non-Volume Preserving Approach to Facial Age-Progression and Age-Invariant Face Recognition}

\author{
	Chi Nhan Duong $^{1,2}$, Kha Gia Quach $^{1,2}$, Khoa Luu $^{2}$ , T. Hoang Ngan Le $^{2}$ and Marios Savvides $^{2}$\\
	$^{1}$ Computer Science and Software Engineering, Concordia University, Montr\'eal, Qu\'ebec, Canada\\
	$^{2}$ CyLab Biometrics Center and the Department of Electrical and Computer Engineering, \\ Carnegie Mellon University, Pittsburgh, PA, USA\\
	{\tt\small \{chinhand, kquach, kluu, thihoanl\}@andrew.cmu.edu, msavvid@ri.cmu.edu}
}

\maketitle

\begin{abstract}
   Modeling the long-term facial aging process is  extremely challenging due to the presence of large and non-linear variations during the face development stages. In order to efficiently address the problem, this work first decomposes the aging process into multiple short-term stages. Then, a novel generative probabilistic model, named Temporal Non-Volume Preserving (TNVP) transformation, is presented to model the facial aging process at each stage. Unlike Generative Adversarial Networks (GANs), which requires an empirical balance threshold, and Restricted Boltzmann Machines (RBM), an intractable model, our proposed TNVP approach guarantees a tractable density function, exact inference and evaluation for embedding the feature transformations between faces in consecutive stages. Our model shows its advantages not only in capturing the non-linear age related variance in each stage but also producing a smooth synthesis in age progression across faces. Our approach can model any face in the wild provided with only four basic landmark points. 
   Moreover, the structure can be transformed into a deep convolutional network while keeping the advantages of probabilistic models with tractable log-likelihood density estimation. 
   Our method is evaluated in both terms of synthesizing age-progressed faces and cross-age face verification and consistently shows the state-of-the-art results in various face aging databases, i.e. FG-NET, MORPH, AginG Faces in the Wild (AGFW), and Cross-Age Celebrity Dataset (CACD).
   A large-scale face verification on Megaface challenge 1 is also performed to further show the advantages of our proposed approach.
\end{abstract}

\section{Introduction}

\begin{figure}[t]
	\centering \includegraphics[width=0.95\columnwidth]{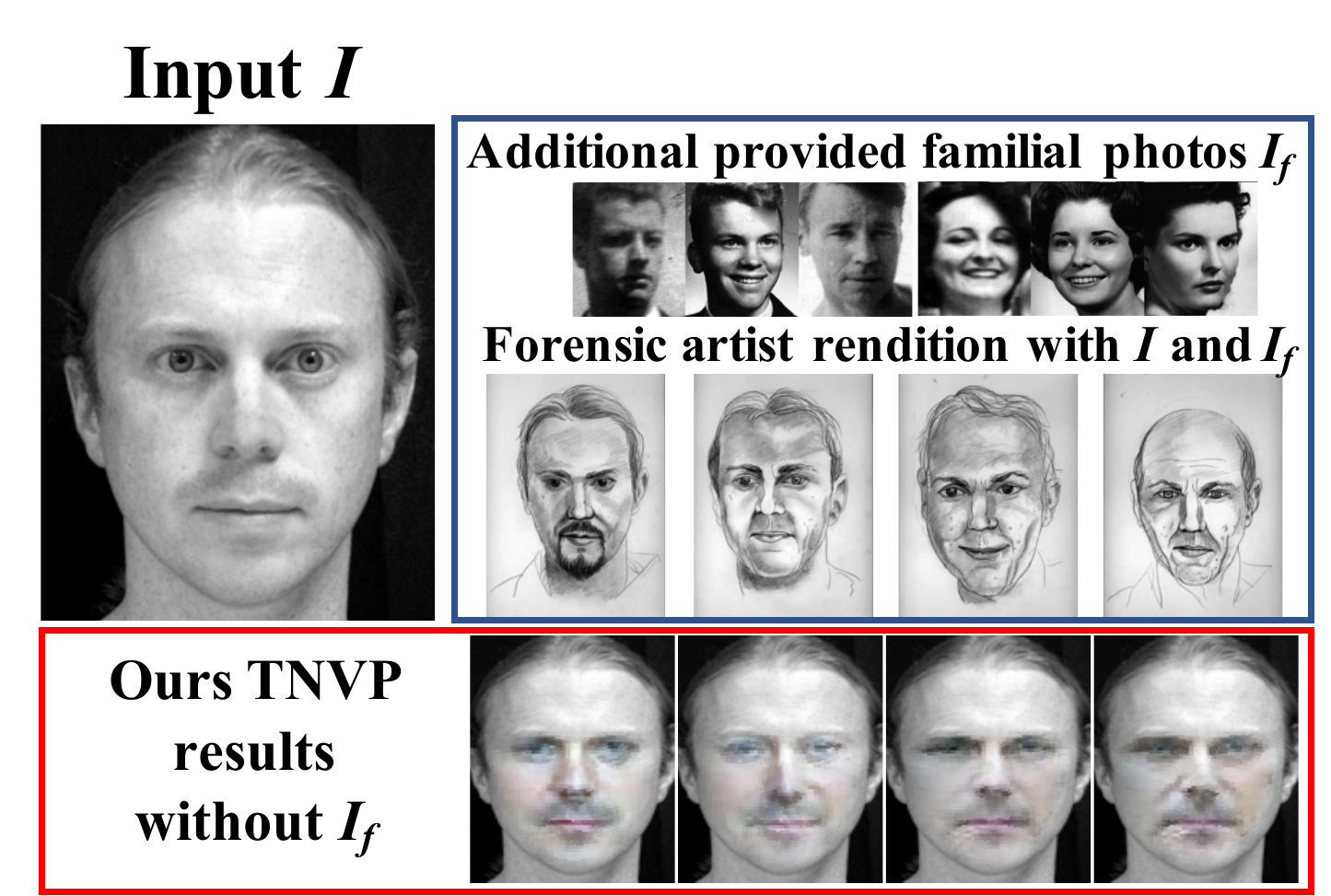}
	\caption{An illustration of age progression from forensic artist and our TNVP model. Given an input $I$ of a subject at 34 years old \cite{patterson2007comparison}, a forensic artist rendered his age-progressed faces at 40s, 50s, 60s and 70s by reference to his familial photos $I_f$. Without using $I_f$, our TNVP can aesthetically produce his age-progressed faces.}
	\label{fig:TNVP_Artist}
\end{figure} 

\begin{table*} [!t]
	\small
	\centering
	\caption{Comparing the properties between our TNVP approach and other age progression methods, where \xmark $ \: $ represents \textit{unknown} or \textit{not directly applicable} properties. 
		Deep learning (DL), Dictionary (DICT), Prototype (PROTO), AGing pattErn Subspace (AGES), Composition (COMP), Probabilistic Graphical Models (PGM), Log-likelihood (LL), Adversarial (ADV) }
	\begin{tabular}{| p{2cm}| p {1.4cm} | p {1.35cm}| p {1.1cm}| p {1.3cm}| p {1.1cm}| p {1.1cm}| p {1.3cm}| p {1.5cm}| p {1.15cm}|}
		\hline
		& Our TNVP & TRBM\cite{Duong_2016_CVPR} & RNN\cite{wang2016recurrent} & acGAN\cite{antipov2017face} & HFA\cite{yang2016face} & CDL\cite{Shu_2015_ICCV} & IAAP\cite{kemelmacher2014illumination} & HAGES\cite{tsai2014human} & AOG\cite{suo2012concatenational} \\ \hline \hline
		\textbf{Model Type} & DL & DL & DL & DL & DICT &  DICT & PROTO & AGES & COMP \\ \hline
		\textbf{Architecture} & PGM+CNN & PGM & CNN & CNN & Bases & Bases & \xmark & \xmark & Graph \\ \hline
		\textbf{Loss Function} & LL & LL & $\ell_2$ & ADV+$\ell_2$ & LL+$\ell_0$ &  $\ell_2+\ell_1$ & \xmark & $\ell_2$ & \xmark \\ \hline
		\hline
		\textbf{Tractable}  &\cmark & \xmark & \cmark & \cmark & \cmark &  \cmark & \xmark & \cmark & \xmark \\ \hline
		\textbf{Non-Linearity} &\cmark    &\cmark
		& \cmark  & \cmark & \xmark & \xmark  & \xmark  & \xmark  & \xmark \\ \hline
	\end{tabular}\label{tab:TenMethodSumm}
\end{table*}

Face age progression is known as the problem of aesthetically predicting individual faces at different ages. 
Aesthetically synthesizing faces of a subject at different development stages is a very challenging task. Human aging is very complicated and differs from one individual to the next. Both \textit{intrinsic factors} such as heredity, gender, and ethnicity, and \textit{extrinsic factors}, i.e. environment and living styles, jointly contribute to this process and create large aging variations between individuals.
As illustrated in Figure \ref{fig:TNVP_Artist}, given a face of a subject at the age of 34 \cite{patterson2007comparison}, a set of closely related family faces has to be provided to a forensic artist as references to generate multiple outputs of his faces at 40’s, 50’s, 60’s, and 70’s. 

In recent years, automatic age progression has become a prominent topic and attracted considerable interest from the computer vision community.
The conventional methods  \cite{geng2007automatic, lanitis2002toward,  patterson2006automatic, suo2012concatenational} simulated face aging by adopting parametric linear models such as Active Appearance Models (AAMs) and 3D Morphable Models (3DMM) to interpret the face geometry and appearance before combining with physical rules or anthropology prior knowledge. Some other approaches \cite{rowland1995manipulating,burt1995perception,kemelmacher2014illumination}
predefined some prototypes and transferred the difference between them to produce age-progressed face images. 
However, since face aging is a non-linear process, these linear models have lots of difficulties 
and the quality of their synthesized results is still limited.
Recently, deep learning based models \cite{Duong_2016_CVPR,wang2016recurrent} have also come into place and produced more plausible results. In \cite{wang2016recurrent}, Recurrent Neural Networks (RNN) are used to model the intermediate states between two consecutive age groups for better aging transition. However, it still has the limitations of producing blurry results by the use of a fixed reconstruction loss function, i.e. $\ell_2$-norm. Meanwhile, with the advantages of graphical models, the Temporal Restricted Boltzmann Machines (TRBM) has shown its potential in the age progression task \cite{Duong_2016_CVPR}. However, its partition function is intractable and needs some  approximations during training process.

\subsection{Contributions of this Work}
This paper presents a novel generative probabilistic model, named Temporal Non-Volume Preserving (TNVP) transformation, for age progression. This modeling approach enjoys the strengths of both probabilistic graphical models to produce better image synthesis quality by avoiding the regular reconstruction loss function, and deep residual networks (ResNet) \cite{He_2016_CVPR} to improve the highly non-linear feature generation. 
The proposed TNVP guarantees a \textit{tractable} log-likelihood density estimation, \textit{exact} inference and evaluation for embedding the feature transformations between faces in consecutive age groups.

In our framework, the long-term face aging is first considered as a composition of short-term stages. Then our TNVP models are constructed to capture the facial aging features transforming between two successive age groups. 
By incorporating the design of ResNet \cite{He_2016_CVPR} based Convolutional Neural Network (CNN) layers in the structure, our TNVP is able to efficiently capture the non-linear facial aging feature related variance. 
In addition, it can be robustly employed on face images in the wild without strict alignments or any complicated preprocessing steps.
Finally, the connections between latent variables of our TNVP can act as ``memory'' and contribute to  produce a smooth age progression between faces while preserving the identity throughout the transitions.

In summary, the novelties of our approach are three-fold. \textbf{(1)} We propose a novel generative probabilistic models with tractable density function to capture the non-linear age variances. \textbf{(2)} The aging transformation can be effectively modeled using our TNVP. Similar to other probabilistic models, our TNVP is more advanced in term of embedding the complex aging process. \textbf{(3)} Unlike previous aging approaches that suffer from a burdensome preprocessing to produce the dense correspondence between faces, our model is able to synthesize realistic faces given any input face in the wild.
Table \ref{tab:TenMethodSumm} compares the properties between our TNVP approach and other age progression methods.

\begin{figure*}[t]
	\centering \includegraphics[width=2\columnwidth]{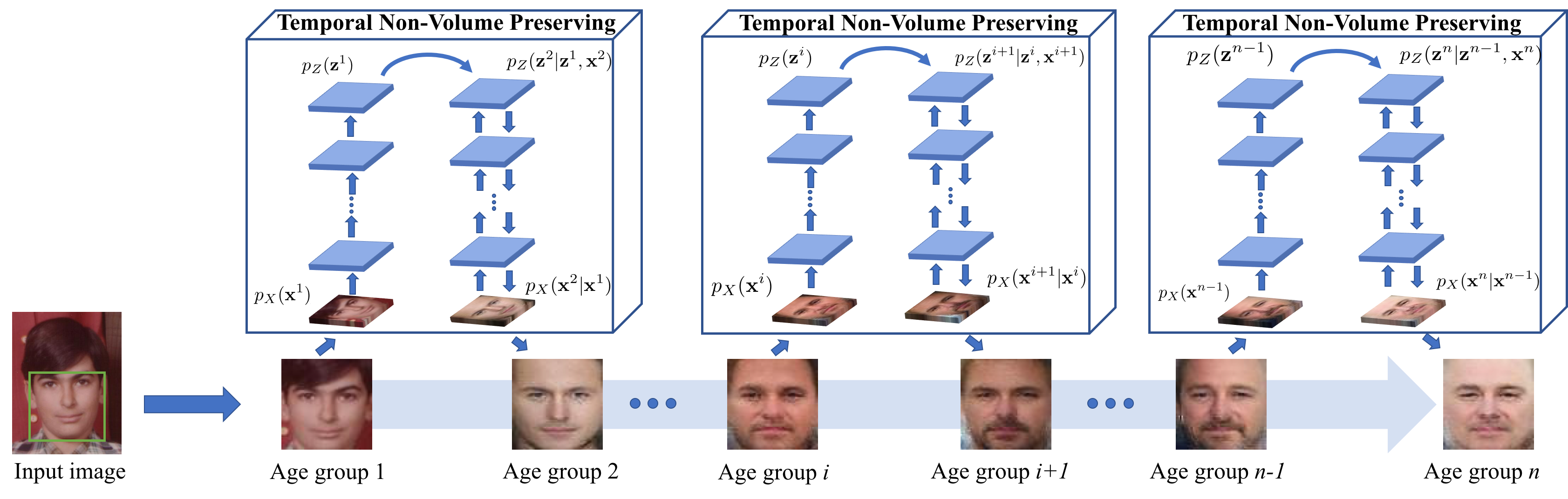}
	\caption{The proposed TNVP based age progression framework. The long-term face aging is decomposed into multiple short-term stages. Then given a face in age group $i$, our TNVP model is applied to synthesize face in the next age group. Each side of our TNVP is designed as a deep ResNet network to efficiently capture the non-linear facial aging features.}
	\label{fig:TNVP_AgeProgression}
\end{figure*}

\section{Related Work}
This section reviews various age progression approaches which can be divided into four groups: \textit{prototyping}, \textit{modeling}, 
\textit{reconstructing}, and 
\textit{deep learning-based approaches}.

\textit{Prototyping approaches} use the age prototypes to synthesize new face images. The average faces of people in the same age group are used as the prototypes \cite{rowland1995manipulating}. The input image can be transformed into the age-progressed face by adding the differences between the prototypes of two age groups \cite{burt1995perception}.
Recently, Kemelmacher-Shlizerman et al. \cite{kemelmacher2014illumination} proposed to construct sharper average prototype faces from a large-scale set of images in combining with subspace alignment and illumination normalization. 

\textit{Modeling-based approaches} represent facial shape and appearance via a set of parameters and model facial aging process via aging functions.
Lanitis et al. \cite{lanitis2002toward} and Pattersons et al. \cite{patterson2006automatic} proposed to use AAMs parameters together with four aging functions for modeling both general and specific aging processes.
Luu et al. \cite{luu2009Automatic} incorporated common facial features of siblings and parents to age progression.	
Geng et al. \cite{geng2007automatic} proposed an AGing pattErn Subspace (AGES) approach to construct a subspace for \textit{aging patterns} as a chronological sequence of face images. Later, Tsai et al. \cite{tsai2014human} improved the stability of AGES by adding subject's characteristics clue. 
Suo et al. \cite{suo2010compositional, suo2012concatenational} modeled a face  using a three-layer And-Or Graph (AOG) of smaller parts, i.e. eyes, nose, mouth, etc. and learned the aging process for each part by applying a Markov chain. 

\textit{Reconstructing-based methods} reconstruct the aging face from the combination of an aging basis in each group. 
Shu et al. \cite{Shu_2015_ICCV} proposed to build aging coupled dictionaries (CDL) to represent personalized aging pattern by preserving personalized facial features. 
Yang et al. \cite{yang2016face} proposed to model person-specific and age-specific factors separately via sparse representation hidden factor analysis (HFA).

Recently, \textit{deep learning-based approaches} are being developed to exploit the power of deep learning methods. 
Duong et al. \cite{Duong_2016_CVPR} employed Temporal Restricted Boltzmann Machines (TRBM) to model the non-linear aging process with geometry constraints 
and spatial DBMs to model a sequence of reference faces and wrinkles of adult faces. 
Similarly, Wang et al. \cite{wang2016recurrent} modeled aging sequences using a recurrent neural network with a two-layer gated recurrent unit (GRU). 
Conditional Generative Adversarial Networks (cGAN) is also applied to synthesize aged images in \cite{antipov2017face}.


\section{Our Proposed Method}
The proposed TNVP age-progression architecture consists of three main steps. (1) Preprocessing; (2) Face variation modeling via mapping functions; and (3) Aging transformation embedding.
With the structure of the mapping function, our TNVP model is tractable and highly non-linear. It is optimized using a log-likelihood objective function that produces sharper age-progressed faces compared to the regular $\ell_2$-norm based reconstruction models.
Figure \ref{fig:TNVP_AgeProgression} illustrates our TNVP-based age progression architecture.

\begin{figure}[t]
	\centering \includegraphics[width=0.9\columnwidth]{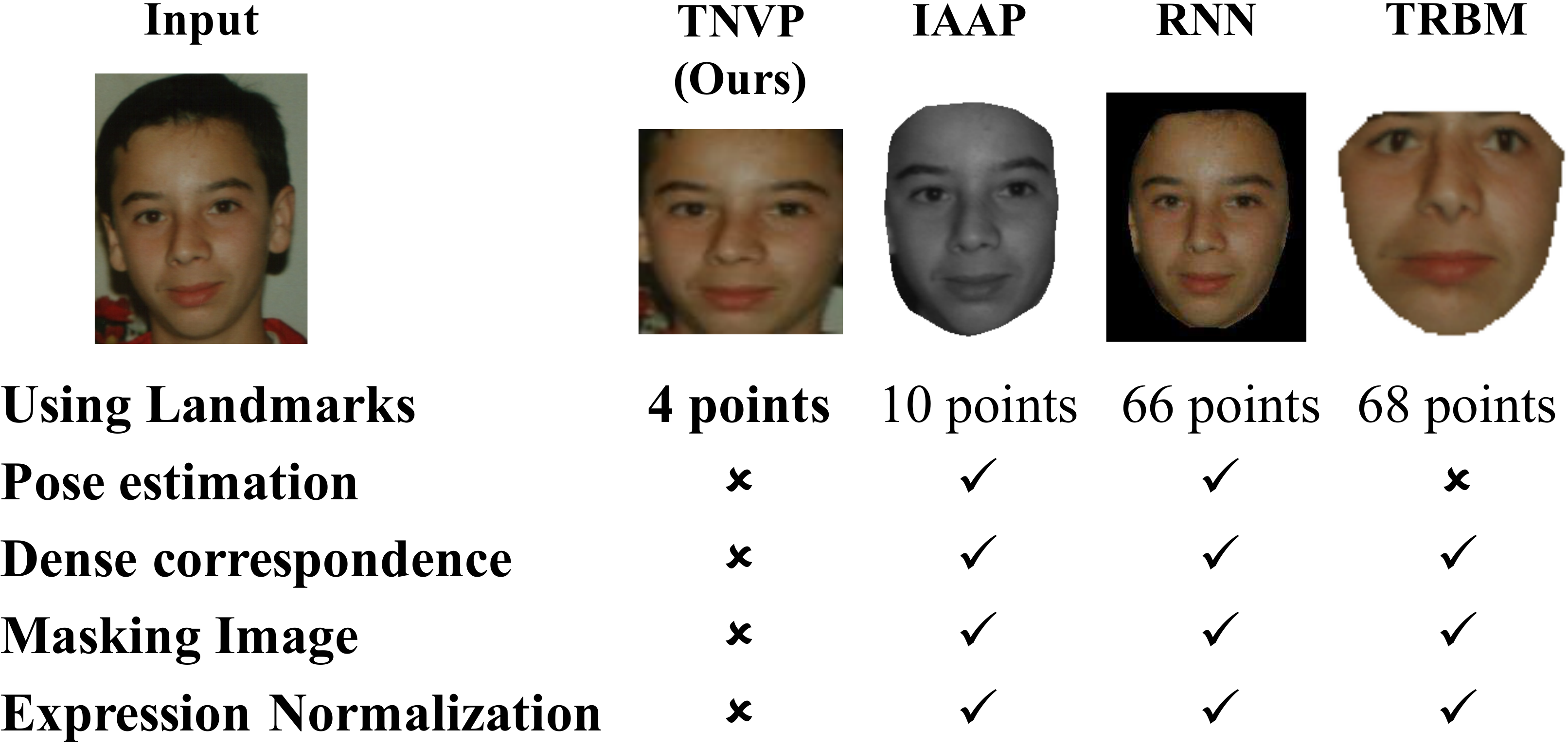}
	\caption{Comparisons between the preprocessing processes of our approach and other aging approaches: IAAP \cite{kemelmacher2014illumination}, RNN based \cite{wang2016recurrent}, and TRBM based \cite{Duong_2016_CVPR} models. Our preprocessing is easy to run, less dependent on the landmarking tools, and efficiently deals with in-the-wild faces. \cmark represents ``included in the preprocessing steps''.
	}
	\label{fig:PreprocessingCompare}
\end{figure}

\subsection{Preprocessing} \label{sec:Preprocessing}
Figure \ref{fig:PreprocessingCompare} compares our preprocessing step with other recent age progression approaches, including Illumination Aware Age Progression (IAAP) \cite{kemelmacher2014illumination}, RNN based \cite{wang2016recurrent}, and TRBM based Age Progression \cite{Duong_2016_CVPR} models.
In those approaches, burdensome face normalization steps are applied to obtain the dense correspondence between faces.
The use of a large number of landmark points makes them highly depend on the stability of landmarking methods that are challenged in the wild conditions. Moreover, masking the faces with a predefined template requires a separate shape adjustment for each age group in later steps.

In our method, given an image, the facial region is simply detected and aligned according to fixed positions of four landmark points, i.e. two eyes and two mouth corners. By avoiding complicated preprocessing steps, our proposed architecture has two advantages. Firstly, a small number of landmark points, i.e. only four points, leverages the dependency to the quality of any landmarking method. Therefore, it helps to increase the robustness of the system. Secondly, parts of the image background are still included, and thus it implicitly embed the shape information during the modeling process. From the experimental results, one can easily notice the change of the face shape when moving from one age group to the next.

\subsection{Face Aging Modeling}
\begin{figure}[t]
	\centering \includegraphics[width=0.95\columnwidth]{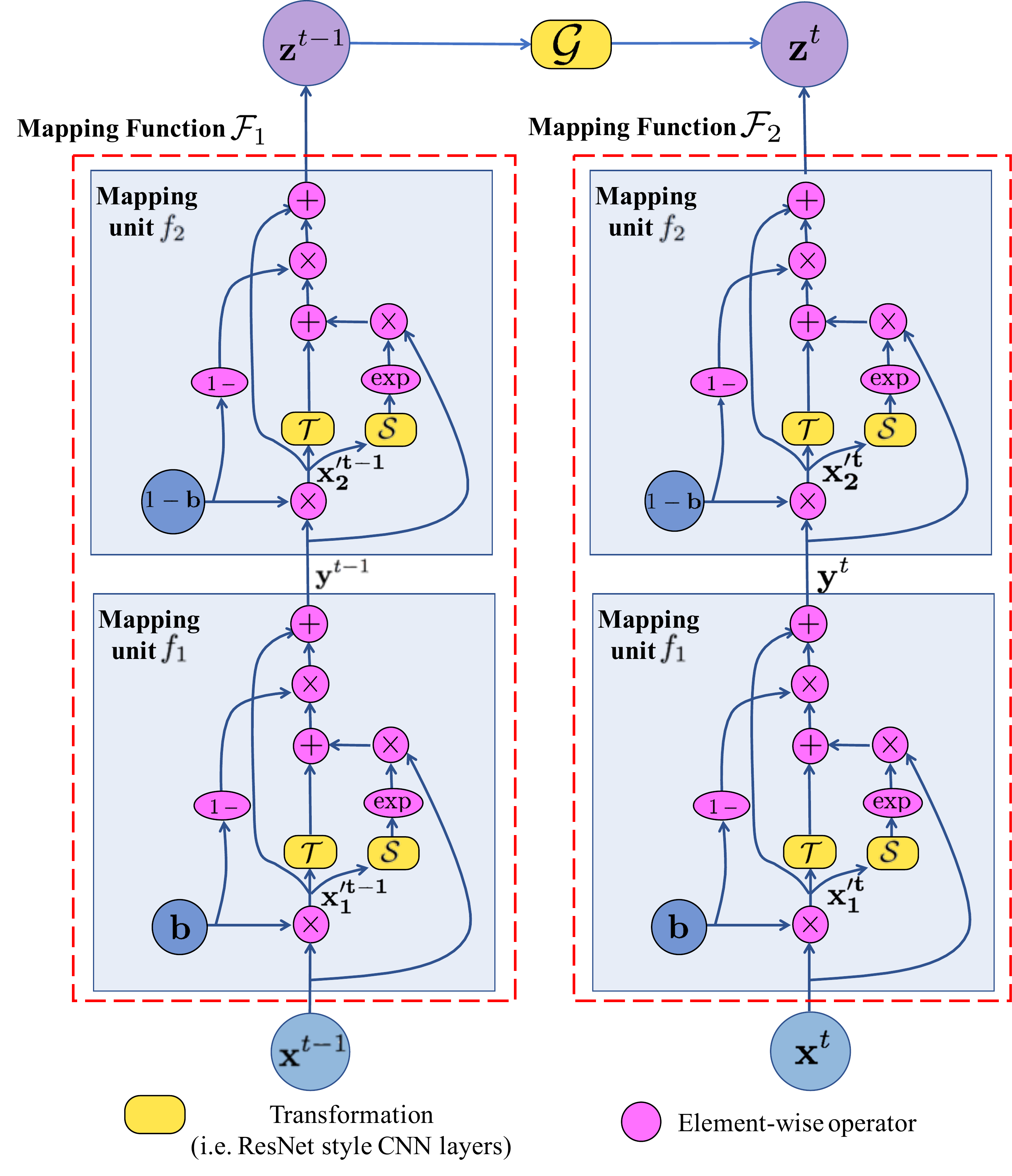}
	\caption{Our proposed TNVP structure with two mapping units. Both transformations $\mathcal{S}$ and $\mathcal{T}$ can be easily formulated as compositions of CNN layers.}
	\label{fig:TNVP_Structure}
\end{figure}

Let $\mathcal{I} \subset \mathbb{R}^D$ be the image domain and $\{\mathbf{x}^t, \mathbf{x}^{t-1}\} \in \mathcal{I}$  be observed variables encoding the texture of face images at age group $t$ and $t-1$, respectively. 
In order to embed the aging transformation between these faces, we first define a bijection mapping function from the image space $\mathcal{I}$ to a latent space $\mathcal{Z}$ and then model the relationship between these latent variables. Formally, let $\mathcal{F}: \mathcal{I} \rightarrow \mathcal{Z}$ define a bijection from an observed variable $\mathbf{x}$ to its corresponding latent variable $\mathbf{z}$ and  $\mathcal{G}:\mathcal{Z} \rightarrow \mathcal{Z}$ be an aging transformation function modeling the relationships between variables in latent space.
As illustrated in Figure \ref{fig:TNVP_Structure}, the relationships between variables are defined as in Eqn. \eqref{eqn:ModelFormulation}.
\begin{equation}
\begin{split}\label{eqn:ModelFormulation}
\mathbf{z}^{t-1}&  = \mathcal{F}_1 (\mathbf{x}^{t-1}; \theta_1)\\
\mathbf{z}^{t} &= \mathcal{H}(\mathbf{z}^{t-1},\mathbf{x}^t; \theta_2, \theta_3)  \\& = \mathcal{G}(\mathbf{z}^{t-1};\theta_3) + \mathcal{F}_2(\mathbf{x}^t;\theta_2)
\end{split}
\end{equation}
where $\mathcal{F}_1, \mathcal{F}_2$ define the bijections of $\mathbf{x}^{t-1}$ and $\mathbf{x}^t$ to their latent variables, respectively. $\mathcal{H}$ denotes the summation of $\mathcal{G}(\mathbf{z}^{t-1};\theta_3)$ and $\mathcal{F}_2(\mathbf{x}^t;\theta_2)$.
$\theta = \{\theta_1,\theta_2,\theta_3 \}$ present the parameters of functions $\mathcal{F}_1, \mathcal{F}_2$ and $\mathcal{G}$, respectively.
Indeed, given a face image in age group $t-1$, the probability density function can be formulated as in Eqn. \eqref{eqn:likelihood}.

\small
\begin{eqnarray}
\begin{split} \label{eqn:likelihood}
p_{X^t}(\mathbf{x}^t|\mathbf{x}^{t-1};\theta)&=p_{X^t}(\mathbf{x}^t|\mathbf{z}^{t-1};\theta)\\
&=p_{Z^t}(\mathbf{z}^t|\mathbf{z}^{t-1};\theta)\left|\frac{\partial \mathcal{H}(\mathbf{z}^{t-1}, \mathbf{x}^t;\theta_2, \theta_3)}{\partial \mathbf{x}^t}\right|\\
&=p_{Z^t}(\mathbf{z}^t|\mathbf{z}^{t-1};\theta)\left|\frac{\partial \mathcal{F}_2(\mathbf{x}^t;\theta_2)}{\partial \mathbf{x}^t}\right|
\end{split}
\end{eqnarray} 
\normalsize
where $p_{X^t}(\mathbf{x}^t|\mathbf{x}^{t-1};\theta)$ and $p_{Z^t}(\mathbf{z}^t|\mathbf{z}^{t-1};\theta)$ are the distribution of $\mathbf{x}^t$ conditional on $\mathbf{x}^{t-1}$ and the distribution of $\mathbf{z}^t$ conditional on $\mathbf{z}^{t-1}$, respectively.
In Eqn. \eqref{eqn:likelihood}, the second equality is obtained using the change of variable formula. $\frac{\partial \mathcal{F}_2(\mathbf{x}^t;\theta_2)}{\partial \mathbf{x}^t}$ is the Jacobian.
Using this formulation, instead of estimating the  density of a sample $\mathbf{x}^t$ conditional on $\mathbf{x}^{t-1}$ directly in the complicated high-dimensional space $\mathcal{I}$, the assigned task can be accomplished by computing the density of its corresponding latent point $\mathbf{z}^t$ given $\mathbf{z}^{t-1}$ associated with the Jacobian determinant $\left|\frac{\partial \mathcal{F}_2(\mathbf{x}^t;\theta_2)}{\partial \mathbf{x}^t}\right|$. 
There are some recent efforts to achieve the tractable inference process via approximations \cite{kingma2013auto} or specific functional forms \cite{dinh2016density,germain2015made, larochelle2011neural}. Section \ref{subsec:mappingfun} introduces a non-linear bijection function that enables the exact and tractable mapping from the image space $\mathcal{I}$ to a latent space $\mathcal{Z}$ where the density of its latent variables can be computed exactly and efficiently. As a result, the density evaluation of the whole model becomes exact and tractable.

\subsection{Mapping function as CNN layers}
\label{subsec:mappingfun}
In general, a bijection function between two high-dimensional domains, i.e. image and latent spaces, usually produces a large Jacobian matrix and is expensive for its determinant computation. Therefore, in order to enable the tractable property for $\mathcal{F}$ with lower computational cost, $\mathcal{F}$ is presented as a composition tractable mapping unit $f$ where each unit can be represented as a combination of several convolutional layers. Then the bijection function $\mathcal{F}$ can be formulated as a deep convolutional neural network.

\subsubsection{Mapping unit} \label{sec:MappingUnit}
\begin{figure}[t]
	\centering \includegraphics[width=1.0\columnwidth]{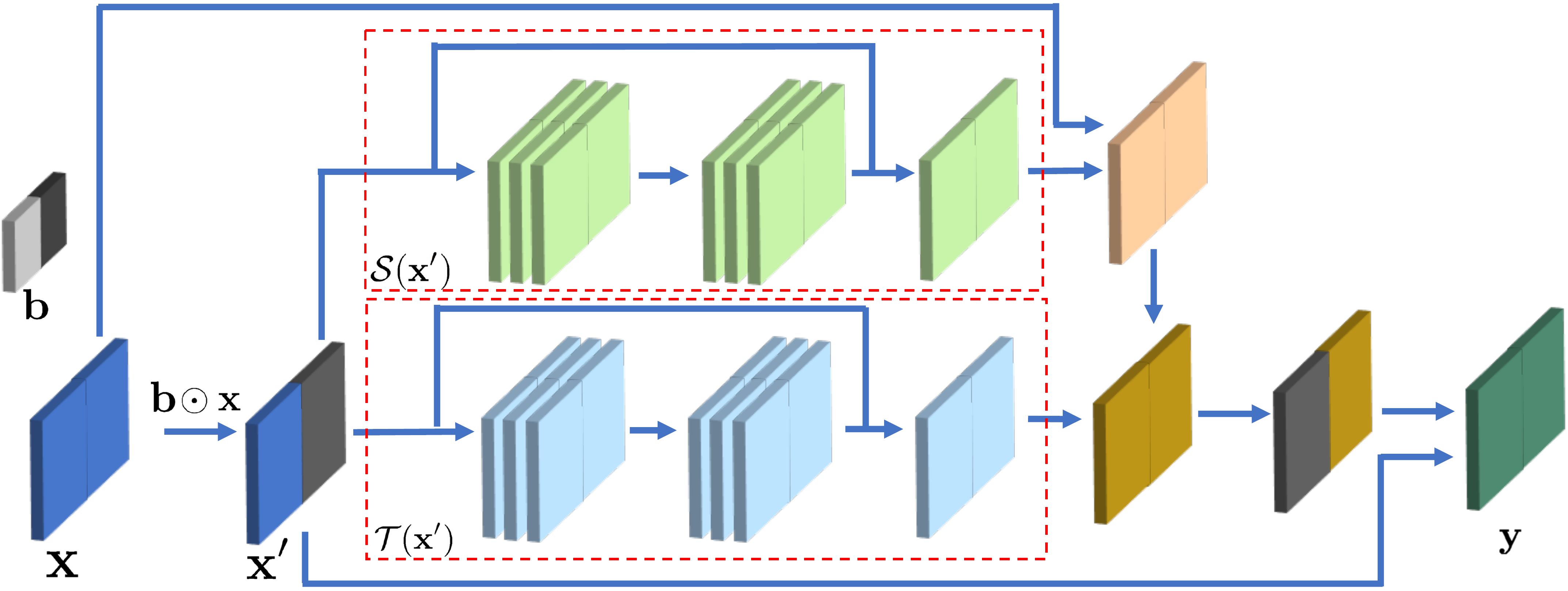}
	\caption{An illustration of mapping unit $f$ whose transformations $\mathcal{S}$ and $\mathcal{T}$ are represented with 1-residual-block CNN network.}  	
	\label{fig:TNVP_MappingUnit}
\end{figure}

Given an input $\mathbf{x}$, a unit $f:\mathbf{x} \rightarrow \mathbf{y}$ defines a mapping between $\mathbf{x}$ to an intermediate latent state $\mathbf{y}$ as in Eqn. \eqref{eqn:maskedConvolution}.
\small
\begin{equation} \label{eqn:maskedConvolution}
\begin{split}
\mathbf{y} = \mathbf{x}' + (1-\mathbf{b}) \odot \left[ \mathbf{x} \odot \exp(\mathcal{S}(\mathbf{x}')) + \mathcal{T}(\mathbf{x}') \right]
\end{split}
\end{equation}
\normalsize
where $\mathbf{x}'=\mathbf{b} \odot \mathbf{x}$; $\odot$ denotes the Hadamard product; $\mathbf{b} = [1,\cdots, 1, 0, \cdots, 0]$ is a binary mask  where the first $d$ elements of $\mathbf{b}$ is set to one and the rest is zero; $\mathcal{S}$ and $\mathcal{T}$ represent the scale and the translation functions, respectively.
The Jacobian of this transformation unit is given by
\begin{equation}
\begin{split}
\frac{\partial f}{\partial \mathbf{x}} &=
\begin{bmatrix}
\frac{\partial \mathbf{y}_{1:d}}{\partial \mathbf{x}_{1:d}}      & \frac{\partial \mathbf{y}_{1:d}}{\partial \mathbf{x}_{d + 1:D}} \\
\frac{\partial \mathbf{y}_{d+1:D}}{\partial \mathbf{x}_{1:d}}       & \frac{\partial \mathbf{y}_{d+1:D}}{\partial \mathbf{x}_{d+1:D}}
\end{bmatrix}
\\
&= 
\begin{bmatrix}
\mathbb{I}_d      & 0 \\
\frac{\partial \mathbf{y}_{d+1:D}}{\partial \mathbf{x}_{1:d}}       & \text{diag}\left(\exp(\mathcal{S}(\mathbf{x}_{1:d}))\right)
\end{bmatrix}
\end{split}
\end{equation}
where $\text{diag}\left(\exp(\mathcal{S}(\mathbf{x}_{1:d}))\right)$ is the diagonal matrix such that $\exp(\mathcal{S}(\mathbf{x}_{1:d}))$ is their diagonal elements. 
This form of $\frac{\partial f}{\partial \mathbf{x}}$ provides two nice properties for the mapping unit $f$.
Firstly, since the Jacobian matrix $\frac{\partial f}{\partial \mathbf{x}}$ is triangular, its determinant can be efficiently computed as,
\small
\begin{equation} \label{eqn:determinant_unit}
\left|\frac{\partial f}{\partial \mathbf{x}}\right| = \prod_j \exp(s_j) =\exp\left( \sum_j s_j\right)
\end{equation}
\normalsize
where $\mathbf{s} = \mathcal{S}(\mathbf{x}_{1:d})$. This property also introduces the tractable feature for $f$.
Secondly, the Jacobian of the two functions $\mathcal{S}$ and $\mathcal{T}$ are not required in the computation of $\left|\frac{\partial f}{\partial \mathbf{x}}\right|$. Therefore, any non-linear function can be chosen for $\mathcal{S}$ and $\mathcal{T}$.
From this property, the functions $\mathcal{S}$ and $\mathcal{T}$  are set up as a composition of CNN layers in ResNet (i.e. residual networks) \cite{He_2016_CVPR}  style with skip connections. This way, high level features can be extracted during the mapping process and improve the generative capability of the proposed model. Figure \ref{fig:TNVP_MappingUnit} illustrates the structure of a mapping unit $f$.
The inverse function $f^{-1}:\mathbf{y} \rightarrow \mathbf{x}$ is also derived as
\begin{equation} \label{eqn:invertMappingUnit}
\small
\begin{split}
\mathbf{x} = & \mathbf{y}' + (1-\mathbf{b}) \odot \left[ (\mathbf{y} - \mathcal{T}(\mathbf{y}')) \odot \exp(-\mathcal{S}(\mathbf{y}'))  \right]
\end{split}
\end{equation}
where $\mathbf{y}'=\mathbf{b} \odot \mathbf{y}$.

\subsubsection{Mapping function} \label{sec:MappingFunction}
The bijection mapping function $\mathcal{F}$ is formulated by composing a sequence of mapping units $\{f_1, f_2, \cdots, f_n\}$.
\begin{equation}
\small
\mathcal{F} = f_1 \circ f_2 \circ \cdots  \circ f_n
\label{eqn:MappingFunc}
\end{equation} 
The Jacobian of $\mathcal{F}$ is no more difficult than its units and still remains tractable.
\begin{equation}
\small
\begin{split}
\frac{\partial \mathcal{F}}{\partial \mathbf{x}} = \frac{\partial f_1}{\partial \mathbf{x}} \cdot \frac{\partial f_2}{\partial f_1} \dots \frac{\partial f_n}{\partial f_{n-1}}
\end{split}
\end{equation}
Similarly, the derivations of its determinant and inverse are
\begin{eqnarray}
\small
\begin{aligned} \label{eqn:DeterminantMappingFunction}
\left|\frac{\partial \mathcal{F}}{\partial \mathbf{x}}\right| &= \left|\frac{\partial f_1}{\partial \mathbf{x}}\right| \cdot \left|\frac{\partial f_2}{\partial f_1}\right| \dots {\left| {\frac{\partial f_n}{\partial f_{n-1}}} \right|}\\
\mathcal{F}^{-1} & = \left(f_1 \circ f_2 \circ \cdots  \circ f_n\right)^{-1} = f_1^{-1} \circ f_2^{-1} \circ \dots  \circ f_n^{-1}
\end{aligned}
\end{eqnarray}
Since each mapping unit leaves part of its input unchanged (i.e. due to the zero-part of the mask $\mathbf{b}$), we alternatively change the binary mask $\mathbf{b}$ to $1-\mathbf{b}$ in the sequence so that every component of $\mathbf{x}$ can be jointed through the mapping process.
As mentioned in the previous section, since each mapping unit is set up as a composition of CNN layers, the bijection $\mathcal{F}$ with the form of Eqn. \eqref{eqn:MappingFunc} becomes a deep convolutional networks that maps its observed variable $\mathbf{x}$ in $\mathcal{I}$ to a latent variable $\mathbf{z}$ in $\mathcal{Z}$.

\subsection{The aging transform embedding}
In the previous section, we present the invertible mapping function $\mathcal{F}$ between a data distribution $p_X$ and a latent distribution $p_Z$. In general, $p_Z$ can be chosen as a prior probability distribution such that it is simple to compute and its latent variable $z$ is easily sampled. In our system, a Gaussian distribution is chosen for $p_Z$, but our proposed model can still work well with any other prior distributions. Since the connections between $\mathbf{z}^{t-1}$ and $\mathbf{z}^t$ embed the relationship between variables of different Gaussian distributions, we further assume that their joint distribution is a Gaussian.
The transformation $\mathcal{G}$ between $\mathbf{z}^{t-1}$ and $\mathbf{z}^t$ is formulated as, 
\begin{equation}
\mathcal{G}(\mathbf{z}^{t-1};\theta_3) = \mathbf{W} \mathbf{z}^{t-1} + \mathbf{b_\mathcal{G}}
\end{equation}
where $\theta_3=\{\mathbf{W, \mathbf{b_\mathcal{G}}}\}$ is the transform parameters representing connecting weights of latent-to-latent interactions and the bias.
From Eqn. \eqref{eqn:ModelFormulation} and Figure \ref{fig:TNVP_Structure}, the latent variable $\mathbf{z}^t$ is computed from two sources: (1) the mapping from observed variable $\mathbf{x}^t$ defined by $\mathcal{F}_2(\mathbf{x}^t; \theta_2)$ and (2) the aging transformation from $\mathbf{z}^{t-1}$ defined by $\mathcal{G}(\mathbf{z}^{t-1};\theta_3)$. The joint distribution $p_{Z^t, Z^{t-1}}(\mathbf{z}^t,\mathbf{z}^{t-1})$ is given by
\small
\begin{eqnarray}
\begin{aligned} 
\label{eqn:JointDistribution}
\mathbf{z}^{t-1} &\sim \mathcal{N}(0,\mathbb{I})\\
\mathcal{F}_2(\mathbf{x}^t,\theta_2) = \bar{\mathbf{z}}^{t} &\sim \mathcal{N}(0,\mathbb{I})\\
p_{Z^t, Z^{t-1}}(\mathbf{z}^t,\mathbf{z}^{t-1}; \theta) & \sim \mathcal{N} 
\left( 
\begin{bmatrix}
\mathbf{b_{\mathcal{G}}}\\
0
\end{bmatrix}
,
\begin{bmatrix}
\mathbf{W}^T\mathbf{W} + \mathbb{I} & \mathbf{W} \\
\mathbf{W} & \mathbb{I}
\end{bmatrix}
\right) 
\end{aligned}
\end{eqnarray}

\normalsize

\begin{figure*}[t]
	\centering \includegraphics[width=1.9\columnwidth]{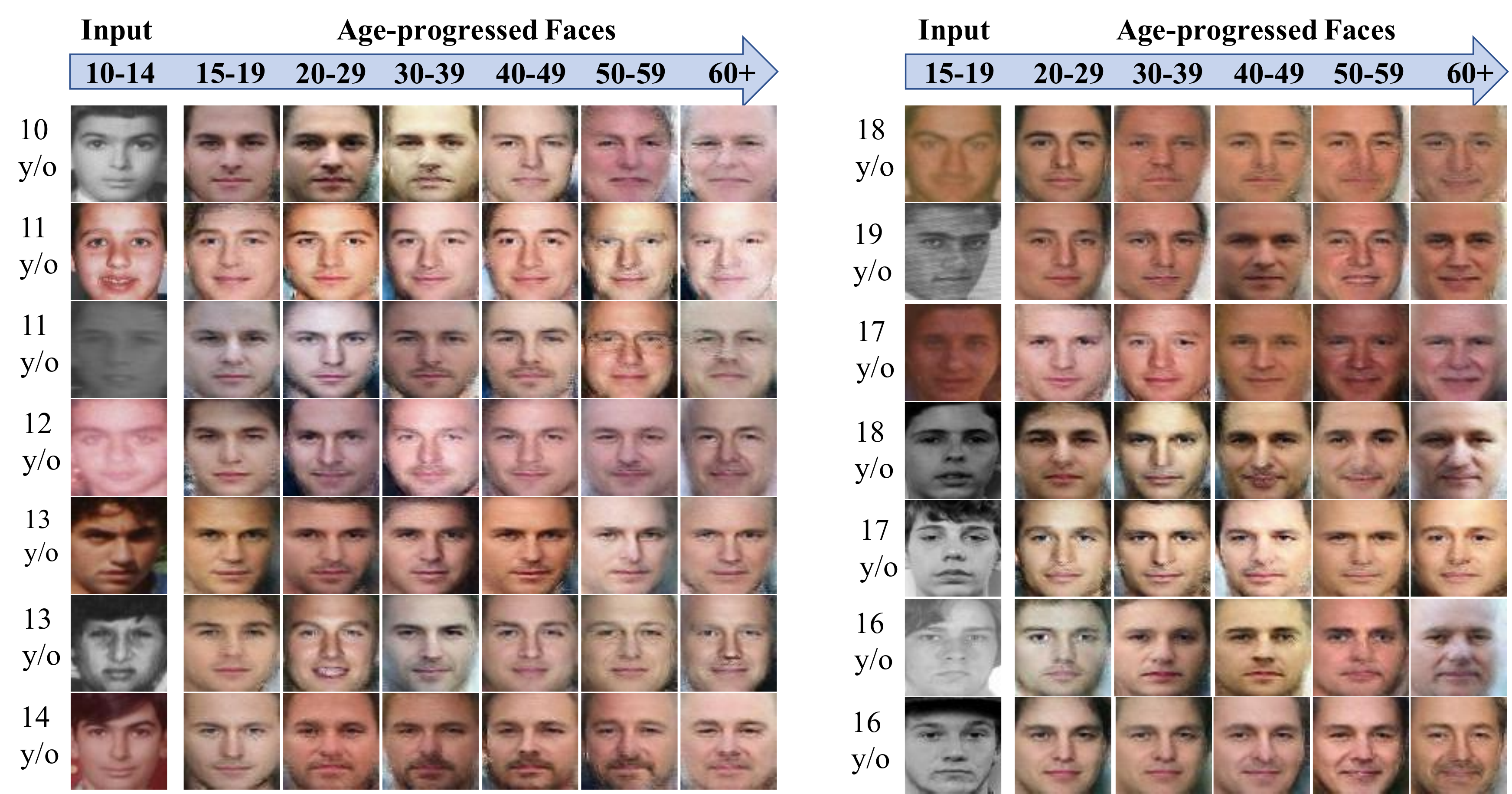}
	\caption{Age Progression Results against FG-NET and MORPH. Given input images, plausible age-progressed faces in different age ranges are automatically synthesized. \textbf{Best viewed in color.}}	
	\label{fig:TNVP_AgeProgressedFaces}
\end{figure*} 

\subsection{Model Learning}
The parameters $\theta=\{\theta_1, \theta_2, \theta_3\}$ of the model are optimized to maximize the log-likelihood:
\begin{equation}
\theta_1^*, \theta_2^*, \theta_3^*=\arg \max_{\theta_1, \theta_2, \theta_3}  \log p_{X^t}(\mathbf{x}^t|\mathbf{x}^{t-1}; \theta_1, \theta_2, \theta_3)
\end{equation}

From Eqn. \eqref{eqn:likelihood}, the log-likelihood can be computed as
\small
\begin{eqnarray}
\begin{split} \nonumber
\log p_{X^t}(\mathbf{x}^t|\mathbf{x}^{t-1}; \theta) = &\log p_{Z^t}(\mathbf{z}^t|\mathbf{z}^{t-1},\theta) + \log \left|\frac{\partial \mathcal{F}_2(\mathbf{x}^t; \theta_2)}{\partial \mathbf{x}^t}\right|\\
= &\log p_{Z^t, Z^{t-1}}(\mathbf{z}^t,\mathbf{z}^{t-1}; \theta)\\
- & \log p_{Z^{t-1}}(\mathbf{z}^{t-1};\theta_1)  + \log \left|\frac{\partial \mathcal{F}_2(\mathbf{x}^t;\theta_2)}{\partial \mathbf{x}^t}\right|
\end{split}
\end{eqnarray} 
\normalsize
where the first two terms are the two density functions and can be computed using Eqn. \eqref{eqn:JointDistribution} while the third term (i.e. the determinant) is obtained using Eqns. \eqref{eqn:DeterminantMappingFunction} and \eqref{eqn:determinant_unit}.
Then the Stochastic Gradient Descent (SGD) algorithm is applied to optimal parameter values.

\subsection{Model Properties}
\textbf{Tractability and Invertibility}: With the specific structure of the bijection $\mathcal{F}$, our proposed graphical model has the capability of modeling arbitrary complex data distributions while keeping the inference process tractable. Furthermore, from Eqns. \eqref{eqn:invertMappingUnit} and \eqref{eqn:DeterminantMappingFunction}, the mapping function is invertible. Therefore, both inference (i.e. mapping from image to latent space) and generation (i.e. from latent to image space)  are exact and efficient. 

\textbf{Flexibility}: as presented in Section \ref{sec:MappingUnit}, our proposed model introduces the freedom of choosing the functions $\mathcal{S}$ and $\mathcal{T}$ for their structures. Therefore, different types of deep learning models can be easily exploited to further improve the generative capability of the proposed TNVP.
In addition, from Eqn. \eqref{eqn:maskedConvolution}, the binary mask $\mathbf{b}$ also provides the flexibility for our model if we consider this as a template during the mapping process. Several masks can be used in different levels of mapping units to fully exploit the structure of the data distribution of the image domain $\mathcal{I}$.

Although our TNVP shares some similar features with RBM and its family such as TRBM, 
the log-likelihood estimation of TNVP is tractable while that in RBM is intractable and requires some approximations during training process. Compared to other methods, our TNVP also shows its advantages in high-quality synthesized faces (by avoiding the $\ell_2$ reconstruction error as in \textit{Variational Autoencoder}) and efficient training process (i.e. avoid maintaining a good balance between generator and discriminator as in case of GANs).

\begin{figure*}[t]
	\centering \includegraphics[width=1.9\columnwidth]{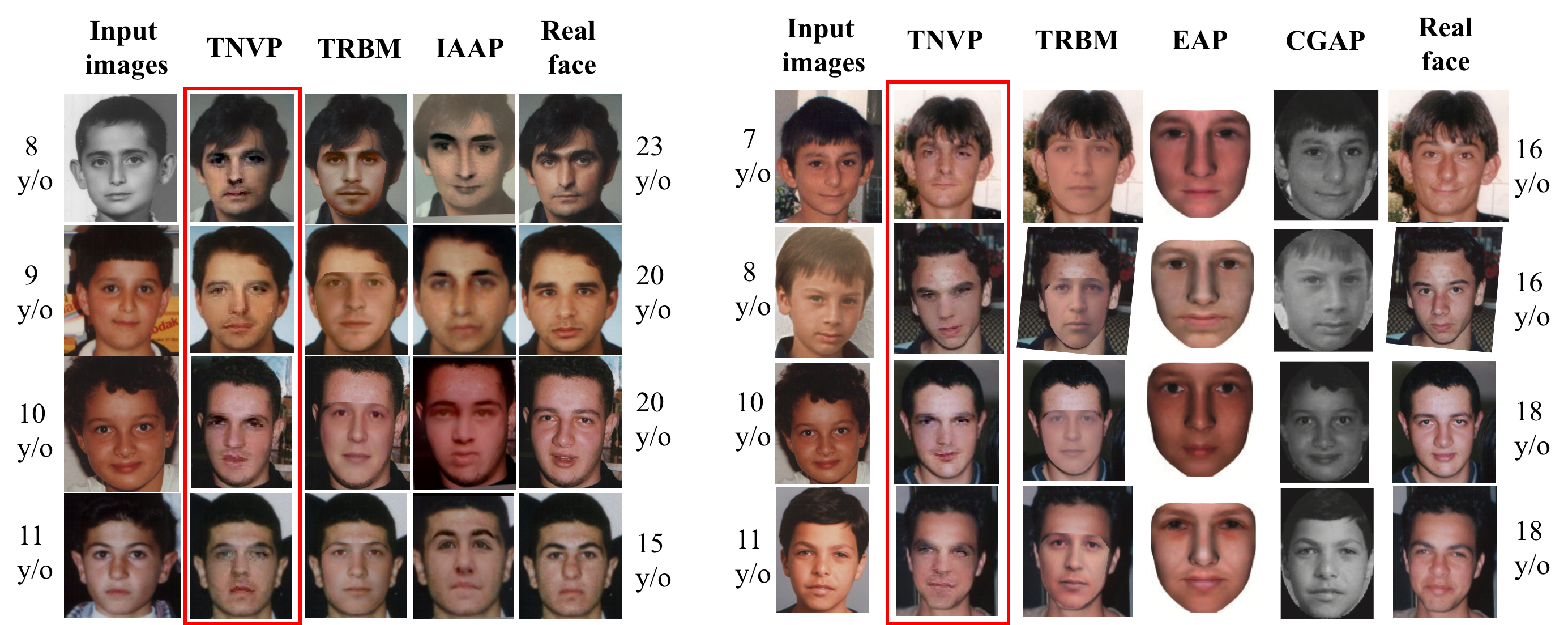}
	\caption{Comparisons between our TNVP against other approaches: IAAP \cite{kemelmacher2014illumination}, TRBM-based \cite{Duong_2016_CVPR}, Exemplar based (EAP) \cite{shen2011exemplar}, and Craniofacial Growth (CGAP) \cite{ramanathan2006modeling} models. \textbf{Best viewed in color.}}	
	\label{fig:TNVP_Comparison}
\end{figure*} 

\section{Experimental Results}
\subsection{Databases}
We train our TNVP system using AginG Faces in the Wild (AGFW) \cite{Duong_2016_CVPR} and  a subset of the Cross-Age Celebrity Dataset (CACD) \cite{chen14cross}. Two other public aging databases, i.e. FG-NET \cite{fgNetData} and MORPH \cite{ricanek2006morph}, are used for testing.

\textbf{AginG Faces in the Wild (AGFW)}: consists of 18,685 images that covers faces from 10 to 64 years old. On average, after dividing into 11 age groups with the span of 5 years, each group contains 1700 images. 

\textbf{Cross-Age Celebrity Dataset (CACD)} is a large-scale dataset with 163446 images of 2000 celebrities. The age range is from 14 to 62 years old. 

\textbf{FG-NET} is a common aging database that consists of 1002 images of 82 subjects and has the age range from 0 to 69. Each face is manually annotated with 68 landmarks.

\textbf{MORPH} includes two albums, i.e. MORPH-I and MORPH-II. The former consists of 1690 images of 515 subjects and the latter provides a set of 55134 photos from 13000 subjects. We use MORPH-I for our experiments.

\subsection{Implementation details}
In order to train our TNVP age progression model, we first select a subset of 572 celebrities from CACD as in the training protocol of \cite{Duong_2016_CVPR}. 
All images of these subjects are then classified into 11 age groups ranging from 10 to 65 with the age span of 5 years. 
Next, the aging sequences for each subject are constructed by collecting and combining all image pairs that cover two successive age groups of that subject. 
This process results in 6437 training sequences. 
All training images from these sequences and the AGFW dataset are then preprocessed as presented in Section \ref{sec:Preprocessing}.
After that, a two-step training process is applied to train our TNVP age progression model.
In the first step, using faces from AGFW, all mapping functions (i.e. $\mathcal{F}_1, \mathcal{F}_2$) are pretrained to obtain the capability of face interpretation and high-level feature extraction. Then in the later step, our TNVP model is employed to learn the aging transformation between faces presented in the face sequences.

For the model configuration, the number of units for each mapping function is set to 10. 
In each mapping unit $f_i$, two Residual Networks with rectifier non-linearity and skip connections are set up for the two transformations  $\mathcal{S}$ and $\mathcal{T}$. Each of them contains 2 residual blocks with 32 feature maps. The convolutional filter size is set to $3 \times 3$. 
The training time for TNVP model is 18.75 hours using a machine of Core i7-6700 @3.4GHz CPU, 64.00 GB RAM and a single NVIDIA GTX Titan X GPU and TensorFlow environment. The training batch size is 64.
\begin{figure*}[!t]
	\centering 
	\begin{subfigure}{0.6\columnwidth}
		\centering 
		\includegraphics[width=1.03\columnwidth]{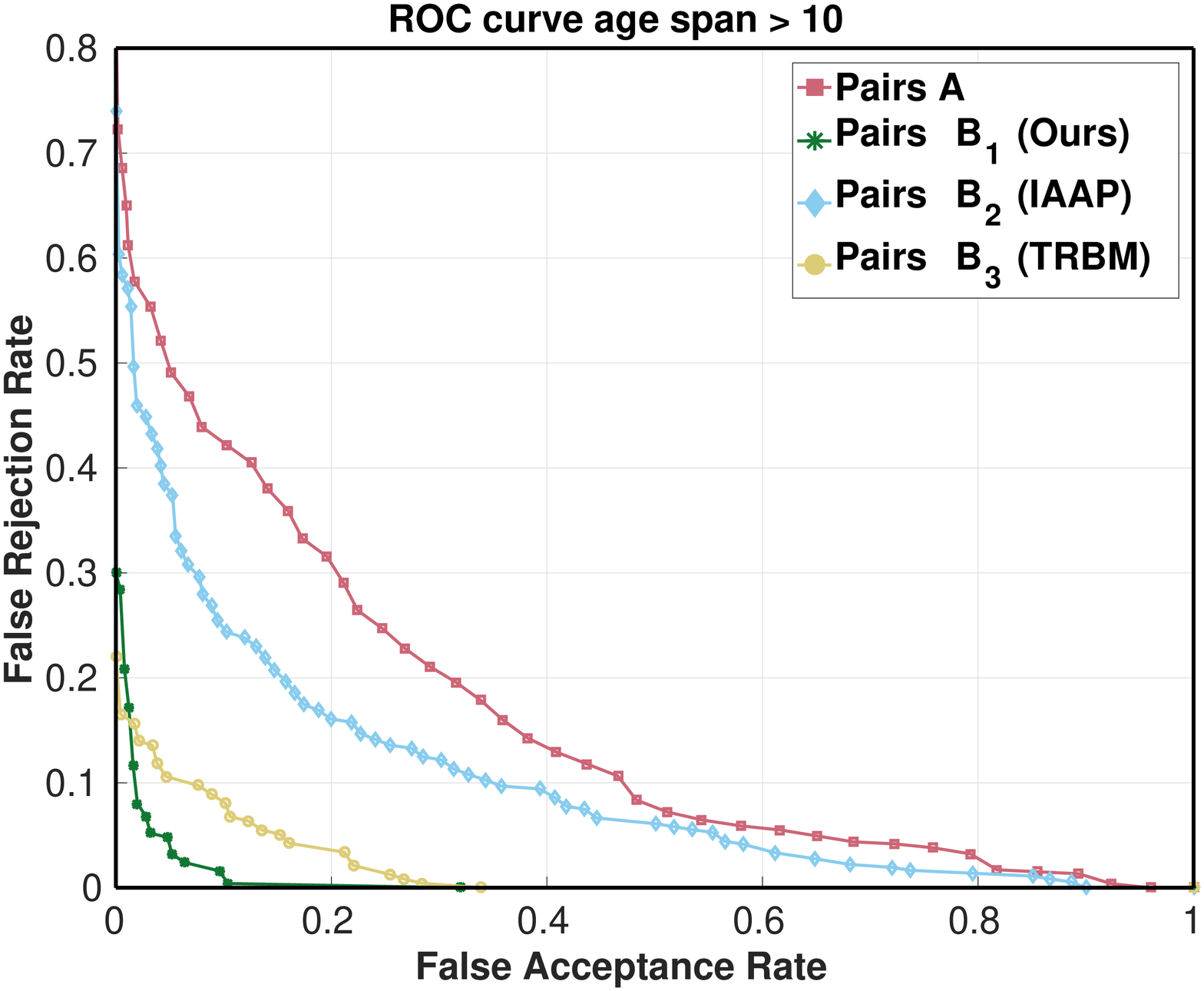}
		\caption{ROC curves of FGNET pairs}
		\label{fig:FG_ROC_Small}
	\end{subfigure} %
	\begin{subfigure}{0.62\columnwidth}
		\centering 
		\includegraphics[width=0.88\columnwidth]{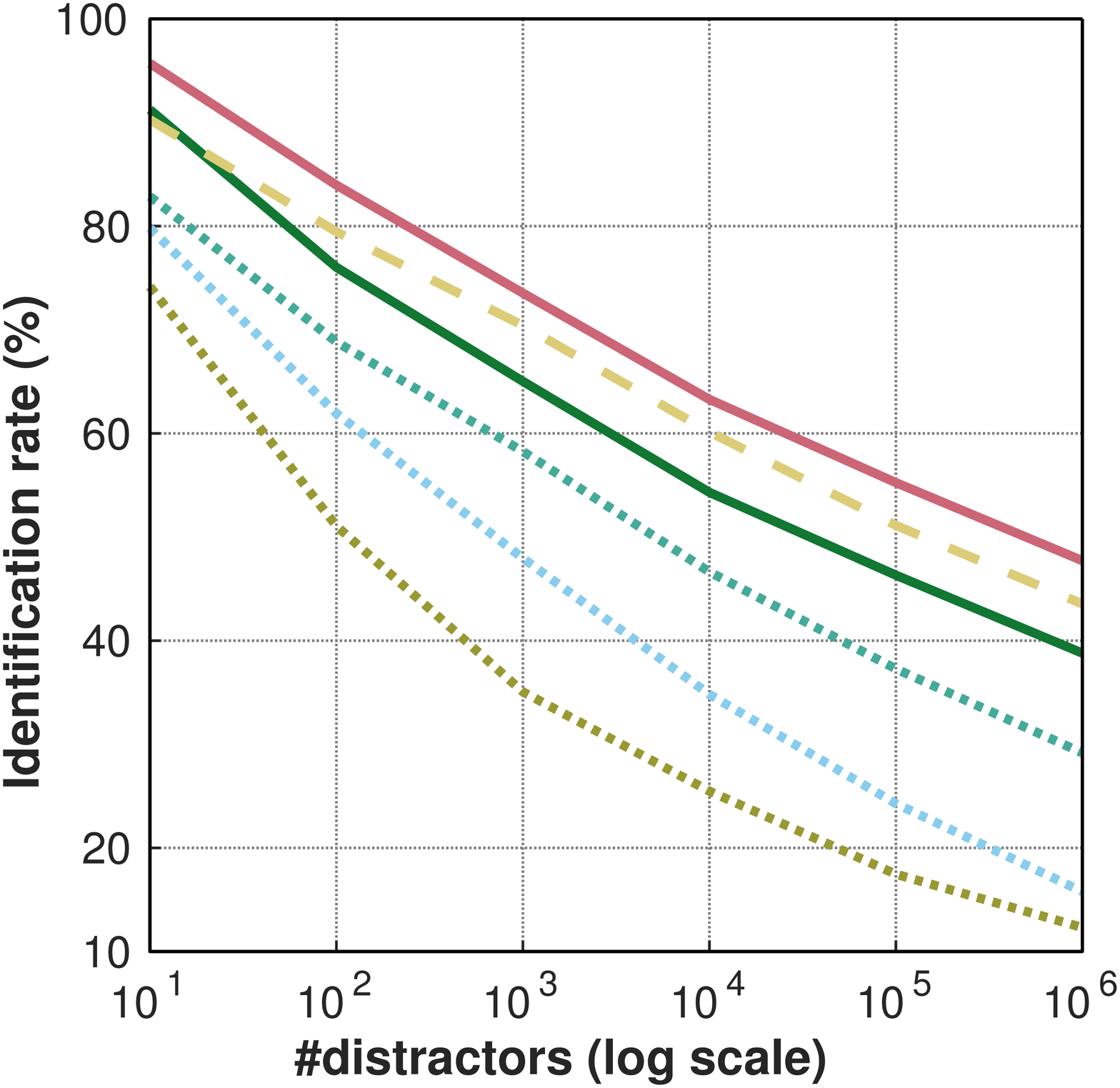}
		\caption{CMC curves on MegaFace}
		\label{fig:megaface_CMC}
	\end{subfigure} %
	\begin{subfigure}{0.8\columnwidth}
		\centering
		\includegraphics[width=1.02\columnwidth]{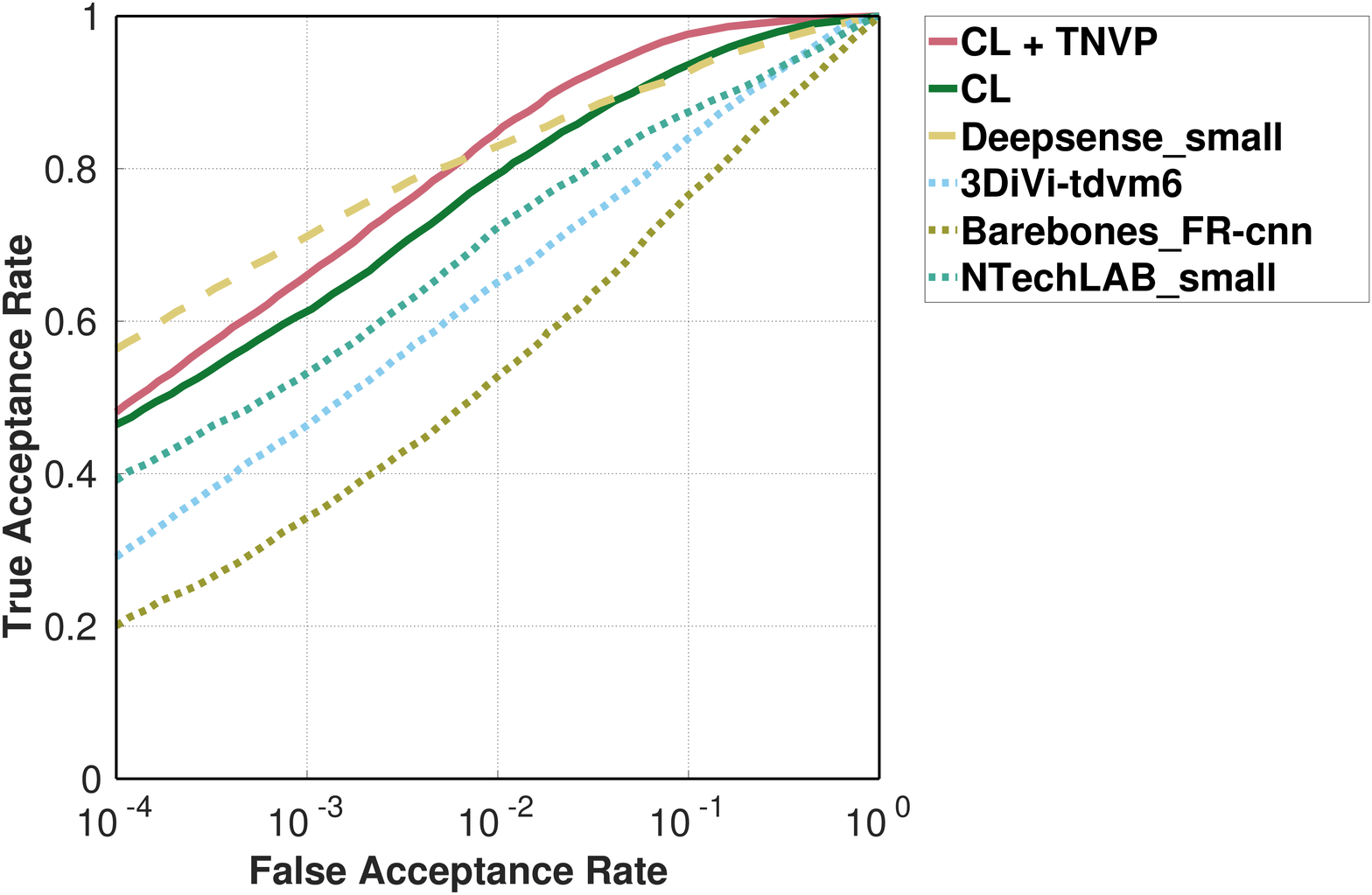}	
		\caption{ROC curves on MegaFace}
		\label{fig:megaface_ROC}
	\end{subfigure} %
	\caption{  
		From left to right: (a) ROC curves of face verification from 1052 pairs synthesized from different age progression methods; (b) ROC and (c) CMC curves of different face matching methods and the improvement of CL method using our age-progressed faces (under the protocol of MegaFace challenge 1).}
	\label{fig:megaface}
\end{figure*}
\subsection{Age Progression}

After training, our TNVP age progression system is applied to all faces over 10 years old from FG-NET and MORPH.
As illustrated in Figure \ref{fig:TNVP_AgeProgressedFaces}, given input faces at different ages, our TNVP is able to synthesize realistic age-progressed faces in different age ranges. Notice that none of the images in FG-NET or MORPH is presented in the training data.
From these results, one can easily see that our TNVP not only efficiently embed the specific aging information of each age group to the input faces but also robustly handles in-the-wild variations such as expressions, illumination, and poses. 
Particularly, beards and wrinkles naturally appear in the age-progressed faces around the ages of 30-49 and over 50, respectively. The face shape is also implicitly handled in our model and changes according to different individuals and age groups.
Moreover, by avoiding the $\ell_2$ reconstruction loss and taking the advantages of maximizing log-likelihood, sharper synthesized results with aging details are produced by our proposed model.
We compare our synthesized results with other recent age progression works whose results are publicly available such as  IAAP \cite{kemelmacher2014illumination}, TRBM-based model \cite{Duong_2016_CVPR} in Figure \ref{fig:TNVP_Comparison}. The real faces of the subjects at target ages are provided for reference. 
Other approaches, i.e. Exemplar based Age Progression (EAP) \cite{shen2011exemplar} and Craniofacial Growth (CGAP) model \cite{ramanathan2006modeling},  are also included for further comparisons. 
Notice that since our TNVP model is trained using the faces ranging from 10 to 64 years old, we choose the ones with ages close to 10 years old during the comparison. These results again show the advantages of our TNVP model in term of efficiently handling the non-linear variations and aging embedding.

\subsection{Age-Invariant face verification}

In this experiment, we validate the effectiveness of our TNVP model by showing the performance gain for cross-age face verification using our age-progressed faces. In both testing protocols, i.e. small-scale with images pairs from FG-NET and large-scale benchmark on Megaface Challenge 1, our aged faces have show that they can provide $\textbf{10 - 40}\%$ improvement on top of the face recognition model without re-train it on cross-age databases. 
We employ the deep face recognition model \cite{wen2016discriminative}, named Center Loss (CL), which is among the state-of-the-art for this experiment. 

Under the \textit{small-scale protocol}, in FG-NET database, we randomly pick 1052 image pairs with the age gap larger than 10 years of either the same or different person. This set is denoted as \textbf{A} consisting of a positive list of 526 image pairs of the same person and a negative list of 526 image pairs of two different subjects.
From each image pair of set  \textbf{A}, using the face with younger age, we synthesize an age-progressed face image at the age of the older one using our proposed TNVP model. This forms a new matching pair, i.e. the aged face vs. the original face at older age. Applying this process for all pairs of set  \textbf{A}, we obtain a new set denoted as set $\mathbf{{B}_1}$. 
To compare with IAAP \cite{tsai2014human} and TRBM \cite{Duong_2016_CVPR} methods, we also construct two other sets of image pairs similarly and denote them as set $\mathbf{{B}_2}$ and $\mathbf{{B}_3}$, respectively. 
Then, the False Rejection Rate-False Acceptance Rate (FRR-FAR) is computed and plotted under the Receiver Operating Characteristic (ROC) curves for all methods (Fig. \ref{fig:FG_ROC_Small}).
Our method achieves an improvement  of $\textbf{30}\%$ on matching performance over the original pair (set \textbf{A}) while IAAP and TRBM slightly increase the rates. 

In addition, the advantages of our model is also experimented on the \textit{large-scale Megaface} \cite{kemelmacher2016megaface} challenge 1 with the FGNET test set. Practical face recognition models should achieve high performance against having gallery set of millions of distractors and probe set of people at various ages. 
In this testing, 4 billion pairs are generated between the probe and gallery sets where the gallery includes one million distractors.
Thus, only improvements on Rank-1 identification rate with one million distractors gallery and verification rate at low FAR are meaningful \cite{kemelmacher2016megaface}.

Table \ref{tb:megaface_test} shows the results of rank-1 identification results with one million distractors. As shown in Fig. \ref{fig:megaface_CMC}, Rank-1 identification accuracy drops when the number of distractors increases.
The TAR-FAR and ROC curves\footnote{The results of other methods are provided in MegaFace website.} are presented in Fig. \ref{fig:megaface_ROC}.
From these results, using our aged face images not only improve face verification results by $\textbf{10}\%$ compared to the original model in \cite{wen2016discriminative} but also outperform most of the models trained with a small training set. The model from DeepSense achieves the best performance under the cross-age training set while the original model trained solely on CASIA WebFace dataset \cite{yi2014learning} having $ < 0.49$M images with no cross-age information. We achieve better performance on top of this original model by simply synthesizing aged images without re-training.

\begin{table}[!t]
	\small
	\centering
	\caption{Rank-1 Identification Accuracy with one million Distractors (MegaFace Challenge 1 - FGNET). Protocol ``small" means $\leq$0.5M images trained. ``Cross-age" means trained with cross-age faces (e.g. in CACD, MORPH, etc.). Center Loss (CL) model in \cite{wen2016discriminative}}
	\label{tb:megaface_test} 
	\begin{tabular}{|l|c|c|c|}
		\hline
		\textbf{Methods} & \textbf{Protocol} & \textbf{Cross-age} & \textbf{Accuracy}\\  \hline \hline
		Barebones\_FR & Small & Y & 7.136 \% \\
		3DiVi & Small & Y & 15.78 \% \\
		NTechLAB & Small & Y & 29.168 \% \\
		DeepSense & Small & Y & 43.54 \% \\
		\hline \hline
		CL \cite{wen2016discriminative} & Small & \textbf{N} & 38.79\% \\
		\textbf{CL + TNVP} & Small & \textbf{N} & \textbf{47.72}\% \\
		\hline
	\end{tabular}
\end{table}

\section{Conclusion}
This paper has presented a novel generative probabilistic model with a tractable density function for age progression. The model inherits the strengths of both probabilistic graphical model and recent advances of ResNet. The non-linear age-related variance
and the aging transformation between age groups are efficiently captured.
Given the log-likelihood objective function, high-quality age-progressed faces can be produced. In addition to a simple preprocessing step, geometric constraints are implicitly embedded during the learning process.
The evaluations in both quality of synthesized faces and cross-age verification showed the robustness of our TNVP.

{\small
\bibliographystyle{ieee}
\bibliography{egbib}

\begin{thebibliography}{10}\itemsep=-1pt

\bibitem{fgNetData}
{\em FG-NET Aging Database}.
\newblock http://www.fgnet.rsunit.com.

\bibitem{antipov2017face}
G.~Antipov, M.~Baccouche, and J.-L. Dugelay.
\newblock Face aging with conditional generative adversarial networks.
\newblock {\em arXiv preprint arXiv:1702.01983}, 2017.

\bibitem{burt1995perception}
D.~M. Burt and D.~I. Perrett.
\newblock Perception of age in adult caucasian male faces: Computer graphic
  manipulation of shape and colour information.
\newblock {\em Proceedings of the Royal Society of London B: Biological
  Sciences}, 259(1355):137--143, 1995.

\bibitem{chen14cross}
B.-C. Chen, C.-S. Chen, and W.~H. Hsu.
\newblock Cross-age reference coding for age-invariant face recognition and
  retrieval.
\newblock In {\em ECCV}, 2014.

\bibitem{dinh2016density}
L.~Dinh, J.~Sohl-Dickstein, and S.~Bengio.
\newblock Density estimation using real nvp.
\newblock {\em arXiv preprint arXiv:1605.08803}, 2016.

\bibitem{geng2007automatic}
X.~Geng, Z.-H. Zhou, and K.~Smith-Miles.
\newblock Automatic age estimation based on facial aging patterns.
\newblock {\em PAMI}, 29(12):2234--2240, 2007.

\bibitem{germain2015made}
M.~Germain, K.~Gregor, I.~Murray, and H.~Larochelle.
\newblock Made: Masked autoencoder for distribution estimation.
\newblock In {\em ICML}, pages 881--889, 2015.

\bibitem{He_2016_CVPR}
K.~He, X.~Zhang, S.~Ren, and J.~Sun.
\newblock Deep residual learning for image recognition.
\newblock In {\em CVPR}, June 2016.

\bibitem{kemelmacher2016megaface}
I.~Kemelmacher-Shlizerman, S.~M. Seitz, D.~Miller, and E.~Brossard.
\newblock The megaface benchmark: 1 million faces for recognition at scale.
\newblock In {\em CVPR}, 2016.

\bibitem{kemelmacher2014illumination}
I.~Kemelmacher-Shlizerman, S.~Suwajanakorn, and S.~M. Seitz.
\newblock Illumination-aware age progression.
\newblock In {\em CVPR}, pages 3334--3341. IEEE, 2014.

\bibitem{kingma2013auto}
D.~P. Kingma and M.~Welling.
\newblock Auto-encoding variational bayes.
\newblock {\em arXiv preprint arXiv:1312.6114}, 2013.

\bibitem{lanitis2002toward}
A.~Lanitis, C.~J. Taylor, and T.~F. Cootes.
\newblock Toward automatic simulation of aging effects on face images.
\newblock {\em PAMI}, 24(4):442--455, 2002.

\bibitem{larochelle2011neural}
H.~Larochelle and I.~Murray.
\newblock The neural autoregressive distribution estimator.
\newblock In {\em AISTATS}, volume~1, page~2, 2011.

\bibitem{luu2009Automatic}
K.~Luu, C.~Suen, T.~Bui, and J.~K. Ricanek.
\newblock Automatic child-face age-progression based on heritability factors of
  familial faces.
\newblock In {\em BIdS}, pages 1--6. IEEE, 2009.

\bibitem{Duong_2016_CVPR}
C.~Nhan~Duong, K.~Luu, K.~Gia~Quach, and T.~D. Bui.
\newblock Longitudinal face modeling via temporal deep restricted boltzmann
  machines.
\newblock In {\em CVPR}, June 2016.

\bibitem{patterson2006automatic}
E.~Patterson, K.~Ricanek, M.~Albert, and E.~Boone.
\newblock Automatic representation of adult aging in facial images.
\newblock In {\em Proc. IASTED Int’l Conf. Visualization, Imaging, and Image
  Processing}, pages 171--176, 2006.

\bibitem{patterson2007comparison}
E.~Patterson, A.~Sethuram, M.~Albert, and K.~Ricanek.
\newblock Comparison of synthetic face aging to age progression by forensic
  sketch artist.
\newblock In {\em IASTED International Conference on Visualization, Imaging,
  and Image Processing, Palma de Mallorca, Spain}, 2007.

\bibitem{ramanathan2006modeling}
N.~Ramanathan and R.~Chellappa.
\newblock Modeling age progression in young faces.
\newblock In {\em CVPR}, volume~1, pages 387--394. IEEE, 2006.

\bibitem{ricanek2006morph}
K.~Ricanek~Jr and T.~Tesafaye.
\newblock Morph: A longitudinal image database of normal adult age-progression.
\newblock In {\em FGR 2006.}, pages 341--345. IEEE, 2006.

\bibitem{rowland1995manipulating}
D.~Rowland, D.~Perrett, et~al.
\newblock Manipulating facial appearance through shape and color.
\newblock {\em Computer Graphics and Applications, IEEE}, 15(5):70--76, 1995.

\bibitem{shen2011exemplar}
C.-T. Shen, W.-H. Lu, S.-W. Shih, and H.-Y.~M. Liao.
\newblock Exemplar-based age progression prediction in children faces.
\newblock In {\em ISM}, pages 123--128. IEEE, 2011.

\bibitem{Shu_2015_ICCV}
X.~Shu, J.~Tang, H.~Lai, L.~Liu, and S.~Yan.
\newblock Personalized age progression with aging dictionary.
\newblock In {\em ICCV}, December 2015.

\bibitem{suo2012concatenational}
J.~Suo, X.~Chen, S.~Shan, W.~Gao, and Q.~Dai.
\newblock A concatenational graph evolution aging model.
\newblock {\em PAMI}, 34(11):2083--2096, 2012.

\bibitem{suo2010compositional}
J.~Suo, S.-C. Zhu, S.~Shan, and X.~Chen.
\newblock A compositional and dynamic model for face aging.
\newblock {\em PAMI}, 32(3):385--401, 2010.

\bibitem{tsai2014human}
M.-H. Tsai, Y.-K. Liao, and I.-C. Lin.
\newblock Human face aging with guided prediction and detail synthesis.
\newblock {\em Multimedia tools and applications}, 72(1):801--824, 2014.

\bibitem{wang2016recurrent}
W.~Wang, Z.~Cui, Y.~Yan, J.~Feng, S.~Yan, X.~Shu, and N.~Sebe.
\newblock Recurrent face aging.
\newblock In {\em CVPR}, pages 2378--2386, 2016.

\bibitem{wen2016discriminative}
Y.~Wen, K.~Zhang, Z.~Li, and Y.~Qiao.
\newblock A discriminative feature learning approach for deep face recognition.
\newblock In {\em ECCV}, pages 499--515. Springer, 2016.

\bibitem{yang2016face}
H.~Yang, D.~Huang, Y.~Wang, H.~Wang, and Y.~Tang.
\newblock Face aging effect simulation using hidden factor analysis joint
  sparse representation.
\newblock {\em TIP}, 25(6):2493--2507, 2016.

\bibitem{yi2014learning}
D.~Yi, Z.~Lei, S.~Liao, and S.~Z. Li.
\newblock Learning face representation from scratch.
\newblock {\em arXiv preprint arXiv:1411.7923}, 2014.

\end{thebibliography}
}

\end{document}